\pdfoutput=1  
\documentclass{article}

\usepackage[numbers]{natbib}
\usepackage[final]{nips_2016}

\usepackage[utf8]{inputenc} 
\usepackage[T1]{fontenc}    
\usepackage{hyperref}       
\usepackage{url}            
\usepackage{booktabs}       
\usepackage{amsfonts}       
\usepackage{nicefrac}       
\usepackage{microtype}      
\usepackage{color}

\usepackage{dirtytalk}

\title{Unknowable Manipulators: \\ Social Network Curator Algorithms}

\author{
    Samuel Albanie \\
    AIMS CDT\thanks{Autonomous Intelligent Machines and Systems, Centre for Doctoral Training}\\
    University of Oxford\\
    \texttt{albanie@robots.ox.ac.uk} \\
    \And
    Hillary Shakespeare \\
    AIMS CDT \\
    University of Oxford \\
    \texttt{hillary@robots.ox.ac.uk} \\
    \And
    Tom Gunter \\
    Engineering Science Department \\
    University of Oxford \\
    \texttt{tgunter@robots.ox.ac.uk} \\
}

\begin{document}

\maketitle

\begin{abstract}
   For a social networking service to acquire and retain users, it must find ways to keep them engaged. By accurately gauging their preferences, it is able to serve them with the subset of available content that maximises revenue for the site.  Without the constraints of an appropriate regulatory framework, we argue that a sufficiently sophisticated  \textit{curator algorithm} tasked with performing this process may choose to explore curation strategies that are detrimental to users.
     In particular, we suggest that such an algorithm is capable of learning to manipulate its users, for several qualitative reasons: 1. Access to vast quantities of user data combined with ongoing breakthroughs in the field of machine learning are leading to powerful but uninterpretable strategies for decision making at scale. 
     2. The availability of an effective feedback mechanism for assessing the short and long term user responses to curation strategies.
     3. Techniques from reinforcement learning have allowed machines to learn automated and highly successful strategies \emph{at an abstract level}, often resulting in non-intuitive yet nonetheless highly appropriate action selection.
    In this work, we consider the form that these strategies for user manipulation might take and scrutinise the role that regulation should play in the design of such systems.
\end{abstract}
\section{Introduction} \label{introduction}

As we approach the year $2020$, access to digital media and services is funnelled through a narrowing oligarchy
 of large technology firms and paid for using those units of barter so favoured by the cash poor millennial generation---fractions of the human attention span and volumes of personal data.  The immense speed and scale at which the domain of social interaction has migrated to the internet has been one of the most striking trends of the last decade. At the heart of this exodus, social networks have emerged as the primary forums of personal, political and commercial discourse \cite{bennett20163}. In such systems, the flow of information depends on the social relationships that link the sub-graphs forming the network and the filtering mechanisms that mediate the interactions along these links. 

To date, the most successful social networks have focused on business models that create value by providing access to a platform which coordinates the sale of advertisements and services to their users (although other revenue sources have been explored \cite{buhler2015big}). For a social network to be financially viable at scale, it must therefore meet two competing demands. It must be sufficiently engaging to acquire and retain new users and it must be effective at advertising products to these users \cite{ballings2015crm}.  In both cases, the central role played by the curation of information in the network is naturally suited to automated approaches \cite{liu2010trends} that can be tuned to maximise the profitability of the site\footnote{For social networks whose business models are  based on advertising, this objective may be maximised through an appropriate proxy, such as the total time a user spends each day interacting with the site.}. Moreover, two key characteristics of internet-based social networks make this filtering task particularly amenable to the use of modern machine learning techniques:  First, access to an unprecedented level of detail corresponding to the historical state of individual users for every previous interaction in which they participated on the network; Second, the availability of sophisticated analytics tools that enable the tracking of user responses to any stimuli they are served by the algorithm.  These analytics provide the system with a powerful feedback mechanism by which it can explore strategies in aid of its optimisation objective. We refer to the collective set of processes used to fulfil this role for a given social network as the \textit{curator algorithm}. 

The action-set of the curator algorithm can be restricted to a single recurring decision for the network: Which subset of available content is to be shown to the user at a given instant? It is clear that the ability of the algorithm to perform this role in an optimal manner is tightly coupled to the information it has access to.   We propose that a \textit{curator algorithm} provided with a large supply of test subjects and an accessible feedback mechanism for evaluating its moves may choose to explore information curation strategies that are detrimental to users.  In particular, we suggest that it may develop sophisticated strategies for manipulating its users as it tries to optimise its given objective.  Moreover, recent trends towards rejecting simpler, interpretable models in favour of more powerful deep architectures that are less amenable to human interpretation make the direct supervision and regulation of the strategies explored by such algorithms extremely difficult. As a consequence, these strategies may be developed without the intention of the network operator.  

The impact of the first generation of social network curator algorithms has attracted significant interest from the research community.  Perhaps the best known hypothesis regarding their usage has been the creation of \say{filter bubbles}. In this phenomenon, users are exposed to an increasingly restricted set of opinions and perspectives by the curation algorithm as it over-exploits its knowledge base about pre-existing user preferences in order to maximise their engagement \citep{pariser2011filter, lazer2015rise}.  Further work has sought to clarify the decisions taken by the algorithms \cite{eslami2015feedvis} and understand the emotional response of users to its application \cite{eslami2015always}.  Related research undertaken by Facebook has emphasised the importance of the individual's choices when determining the extent to which curation influences a user's exposure to challenging views \cite{bakshy2015exposure}. These studies provide a useful context for the effects produced by early attempts at social network curation.  However, in this work we instead focus our attention on the potential consequences of the next generation of viable curator algorithms.

A number of previous works have also explored the potential for forms of Artificial Intelligence to manipulate humans, particularly as a consequence of a predicted intelligence explosion \citep{good1966speculations, vinge1993coming}, an event which is often referred to as the \textit{singularity} \cite{ulamtribute}. The many risks of human manipulation by the resulting \textit{superintelligence} are analysed in detail in \cite{bostrom2014superintelligence}. Previous predictions for the timescale of this event vary, but all consider that if it were to occur, it would require a level of technology that is not yet available \cite{bostrom1998long, kurzweil2005singularity, bostrom2014superintelligence}. In contrast to the threat posed by a superintelligence, we argue that the algorithmic manipulation of humans in social networks is feasible with currently available technology. 

More closely related to our work, the potential for \textit{psychological parasites} (intellectual stimuli that lead to addictions) are identified as a risk associated with the improving capabilities of technology in \cite{heylighen2014return}. These risks are particularly abundant in \textit{mobilsation systems}---persuasive technologies designed to coordinate users towards specific goals \cite{heylighen2013mobilization}. We develop this idea further, arguing that there are specific risks posed by the combination of current machine learning algorithms and access to abundant user data in the social network domain. 

Set to come into force in 2018, the European General Data Protection Regulation \cite{GDPR} introduces a range of measures of significant relevance to industries heavily engaged in the collection and analysis of user data.   Any framework that seeks to provide appropriate regulation for curator algorithms faces a daunting task: it must seek to protect the well-being of the network participants but also strive to protect the ability of the network operators to innovate.  We consider the effect of this legislation in the social network domain and assert that curator algorithms deserve particular attention from regulators.  

When considering the potential avenues for the regulation of curation algorithms, it may prove useful to consider how other industries have approached similar challenges.  In recent years, regulators in the financial industry have been faced with the task of preventing market manipulation by increasingly complicated, algorithmically driven high frequency trading strategies \cite{gomber2006catching}. While some of the proposed regulatory responses are specific to finance (for instance, cancellation taxes which render a number of market manipulation strategies infeasible \cite{prewitt2012high}), the financial industry provides a useful reference point for regulators in the social network domain (see Sec. \ref{regulation} for details). 

In this work we consider the risks and regulation of social network curator algorithms by formulating their task as a \textit{reinforcement learning} problem. Concretely, our first contribution is to determine the risks of an unregulated system by exploring a range of strategies a \textit{curator algorithm} might employ with detrimental effects for users.  Our second contribution is to propose specific strategies for the safe regulation of \textit{curator algorithms} in the context of existing data legislation and to assess their potential effectiveness in this role.

\section{Engagement as a Learning Problem} \label{learning}
We will view the problem of maximising user engagement according to some utility function much as a machine learning researcher working in advertisement might---as a reinforcement learning task \cite{sutton1998reinforcement}.   This framework has been shown to be particularly effective in optimising content selection for social network users \cite{li2010contextual}. 

At a coarse level, a typical reinforcement learning model is built around several core concepts:

\begin{itemize}
    \item A set of states, $S$, which fully encode the system and environment we intend to model. The state for individual users may be modelled as an aggregate of the content presented on-screen and a (partially observed) estimate of the user's `internal' mental state.
    \item A set of possible actions, $A$, which the system can trigger in return for a (possibly delayed) reward ($R$).  Triggering an action \emph{may} also cause a state transition.  In the examples we consider, an action may represent content to a social-network user.
    \item An indication of reward, utility, or long term value for the algorithm ($R$).  It is against this that the operator adapts the policy function, selecting for strategies which maximise this reward.  
    \item A policy function $P : S \times A \rightarrow R (\times S)$. This mapping essentially encodes the strategy which the system pursues in order to maximise reward in the long term horizon.  It is here that external control of the curation algorithm may be exerted to avoid pathological and potentially unethical behaviour.
\end{itemize}

Reinforcement learning anneals on a policy function to maximise the value and therefore the long running utility of a system. It is clear then, that it is this component which determines the sophistication of the user engagement strategy, and therefore it is here that we focus our attention.
At a fundamental level, the policy function does nothing more than provide a mapping from the state-action space through to the scalar value function. 
The sophistication of the strategy is therefore strongly linked to the complexity of the mapping we are able to express, and today deep neural networks are usually chosen as the surrogate for this function. 
These are capable of expressing very complex and non-intuitive functions, as demonstrated by Google's AlphaGo project \cite{silver2016mastering}, where a Go playing policy function was learned which not only outperformed top human players, but did so via a mixed mode of human-like and highly non-intuitive but optimal moves.
Other examples of such behaviour arose when these systems were trained to play video games. In particular, when Google trained a policy for playing the notorious Atari boxer game \cite{mnih2013playing} the system learned to exploit weaknesses in the game design, trapping the opponent in a corner and thereby guaranteeing victory.  
As research continues, we can envisage a world in which these approaches are effectively brought to bear on the \say{game} of maximising network profitability. If governed solely by this utility function, we suggest that equivalent pathologies in human behaviour may be discovered and exploited.

\section{Manipulation Through Curation} \label{manipulation}

In this section we discuss the range of manipulation strategies available to a curator algorithm seeking to optimise the profitability of a social network. In this context, we take manipulation to mean \textit{the art of deliberately influencing a person's behaviour to benefit some objective}. We begin by describing the forms of manipulation that are applicable in the domain of social networks. We then introduce a simple categorisation of the different forms of manipulation and offer examples of the strategies a curator algorithm might develop with detrimental effects for its users.

Manipulation forms a natural component of human interaction and can take many forms, ranging from direct requests to subtle and intentionally hidden signals. A number of previous studies have demonstrated how human behaviour can be influenced with subtle visual and verbal clues  \citep{ernest2011effects, kettle2016behavioral, laakasuo2015emotional, leonard2008richard}. Research into the design of site features has further demonstrated the ability of operators to \say{steer} user behaviour in \cite{anderson2013steering}. Of particular relevance to this work, it has been shown that the emotional states of social network users can be influenced by selectively filtering the content produced by their friends \cite{kramer2014experimental}.  

Influential early work in the field of behavioural psychology determined that animals could be manipulated most effectively if they are rewarded on a variable, unpredictable schedule \cite{skinner1938behavior}. This behaviour has been used profitably by casinos who offer gamblers surprise rewards to keep them hooked to the action in the midst of a losing streak \cite{binkley2004taking}.  Similar ideas have been applied to game design to keep players engaged for longer by unpredictably varying the duration of in-game tasks \cite{zagal2013dark}.  These psychological traits exemplify the kind of in-built behaviours that could be discovered and exploited by a curator algorithm. 

In order to explore the specific forms of strategy available to a curator algorithm we propose a simple categorisation of manipulation. We define a manipulation to be of \textit{first order} if the manipulation is direct and the objective of the manipulator is transparent to the participant. A manipulation is defined to be of \textit{second order} if it is indirect, but the objective remains transparent to the participant.  Further, we consider a manipulation to be of \textit{third order} if it is indirect and the means by which the objective is attained are not transparent to the participant\footnote{Note that the distinction between these categories rests on the difficult assessment of the cognitive abilities of the target \cite{yoshida2008game}. The manipulator may determine that the intention of a given set of behaviours is transparent to a sophisticated target, but not to a simple target.}.

These categories may be illustrated with a simple example. Consider a bar owner wishing to increase drinks sales at their establishment. Each evening, the owner may choose to simply ask customers directly to purchase more drinks. This strategy, corresponding to a first order manipulation, has the benefit of simplicity but may not lead to optimal drinks sales (or indeed the renewal of their bar licence). The owner may instead aim to increase sales with advertisements illustrating the enjoyment of other customers as they refresh themselves with drinks from the bar.  This form of advertising aims to evoke a sense of desire in the customers which may lead indirectly to the purchase of more drinks.  However, the objective of the advert remains transparent to the customer, corresponding to a second order manipulation.  Finally, a shrewd bar owner may employ a third strategy, in which they provide free snacks to customers of the bar. The snacks, however, are heavily salted, and after consuming them the customers find their throats parched and in need of immediate refreshment.  This strategy is both indirect and not transparent to all but the experienced customers, corresponding to a third order manipulation. 

We remark here that access to detailed information about the target plays an important role in the ability to manipulate them.  Should an unethical bar owner overhear sensitive information about the personal life of a customer at the bar they have the potential to pursue further strategies, such as ensuring the customer loses their job so that they are more likely to spend time drinking at the establishment.

We might assume that a curator algorithm seeking to maximise profitability will naturally explore first and second order manipulations as it seeks to advertise products to its user base. Aided by access to detailed user information, it can make powerful inferences about which information should be displayed at each instant. Consider, for example, the marketing of an energy drink.  With the knowledge that a user is a student, that they are awake beyond their usual sleep cycle, that the date of their exams is drawing near and that their online activity shows indications of fatigue, the curator can select an optimal time and context for the display of an advert.  Now imagine a more sophisticated algorithm capable of pursuing third order manipulations. Such an algorithm might choose to display content which had been selected with the specific goal of exhausting the user.  This could be achieved by triggering predictable repeat behaviours gleaned from an in-depth knowledge of their browsing habits. Indeed, over longer time horizons, the curator might determine that an effective method for increasing the sales of energy drinks is the distortion of the user's sleeping patterns.  To take another example, consider a curator algorithm seeking to use information about social groups to increase sales of dating site memberships. While simple manipulations could lead it to present content encouraging individual users to search for partners, it could pursue third order manipulations by intentionally encouraging subsets of social groups to communicate in a manner that excludes other members, actively evoking a feeling of loneliness in the affected party to increase their responsiveness to advertising.  

A recent example of this strategy exploration principle in action can be found in the efforts of a collection of companies seeking to optimise advertising revenue during the United States presidential election in $2016$  \cite{fakenews}. Through simple trial and error, they determined that carefully targeted fake political news stories were extremely effective in maximising click-throughs. Since this strategy was optimising their objective, they doubled down on this approach and produced as much content as possible without regard for its effect on the users of the network.  With the same objective, even a comparatively simple curator algorithm would be capable of developing this strategy.

We note that it is certainly not the case that all strategies pursued by a curator algorithm will be detrimental for users.  Indeed, the energy drinks may give the tired student the boost required to raise their grade, while the previously lonely user may find happiness through their new dating site membership. However, perhaps the most striking aspect of the Atari game-playing algorithm \cite{mnih2013playing} was not that it was capable of surpassing human performance, but rather that it came up with \say{cheat} strategies that human players had not previously considered (e.g. the boxer strategy described in Sec.\ref{learning}).  Similarly, although the manipulation examples described above are simple and interpretable, we suggest that the curator algorithm is capable of developing sophisticated, uninterpretable strategies for manipulating users as they optimise their objective.  By their very nature, such strategies are difficult to predict and therefore difficult to regulate.  It is however an issue that is worthy of consideration if we wish to avoid the discovery of similar \say{cheat} strategies for human manipulation.
\section{Curator Regulation} \label{regulation}

Do social network curator algorithms deserve special attention from regulators?  In Sec. \ref{introduction}, we argued that the risks of higher order manipulations result from providing curator algorithms with three key assets: extensive access to user data, the ability to devise sophisticated strategies (potentially beyond the understanding of human operators) and an effective mechanism for evaluating the effects of its strategies.   In this section, we assess the need for regulation in social network curator algorithms in the context of these three areas. We begin by discussing the General Data Protection Regulation (GDPR) recently introduced by the European Union and its implications for the strategies available to algorithms operating in the social network domain.  Specifically, we examine its ability to safeguard users from higher order manipulations through its requirement of algorithm \textit{interpretability}.  Next, we explore strategies for the specific regulation of reinforcement learning-based curator algorithms and make recommendations for their application.  Finally, we discuss the challenges faced by regulators operating in industries in which algorithm interpretability is often infeasible as a useful reference for regulators considering the problem of curator manipulation.  

As modern social networks develop a global user base, they become subject to a diverse range of national data and privacy regulations \cite{bygrave2014data}, as well as laws governing the \textit{transborder data flows} that occur in the operation of a international organisation \cite{kuner2013transborder}.  Of these, one set of regulations which holds particular significance for the operation of curator algorithms is the \textit{General Data Protection Regulation}, set to come into force across the European Union in 2018 \cite{GDPR}.  Among rules governing the storage and usage of personal data which will apply in social network domain, its creation of a so called \say{right to explanation} \cite{goodman2016eu} has significant consequences for the design of algorithms that operate on personal data.  By requiring that companies performing automated decision making based on personal data must be capable of supplying \say{meaningful information about the logic involved}, the regulation places heavy emphasis on \textit{algorithm interpretability}.  

Algorithm interpretability has been a longstanding topic of interest in machine learning, yielding techniques that modify and extend models to explain their decisions \citep{swartout1983xplain, ridgeway1998interpretable} alongside efforts to improve naturally interpretable algorithms to make them competitive with their opaque counterparts \cite{lou2012intelligible}.  However, while there has been a great deal of research interest in improving the understanding of deep neural networks (using techniques such as random perturbation \cite{olden2002illuminating}, invariance analysis \citep{goodfellow2009measuring, bruna2013invariant, Lenc15}, visualisation \citep{simonyan2013deep, zeiler2014visualizing, mahendran2015understanding} and dimensionality reduction \cite{vellido2012making}), the interpretation of these models remains notoriously challenging.  Consequently, it is not clear whether these models are currently capable of providing the \say{meaningful information} required by the regulation. 

To achieve compliance, curator algorithms may therefore be restricted to a set of simple function classes.  As a result, the potential sophistication of the policy function described in Sec. \ref{learning} would be curtailed and higher order manipulation strategies would be unlikely to arise. However, we note that there are two reasons why this regulation may not be an effective safeguard for social network users.  Firstly, the regulation sets such comprehensive requirements that it may become meaningless in the law books \cite{koops2014trouble}.  Secondly, a lack of consensus on precisely what it means for a model to possess the property of interpretability or be capable of providing \say{meaningful information} makes it extremely difficult to assess the forms of algorithm that would be compliant with the regulation.  Indeed, under certain criteria it has been observed that deep neural networks may be considered no less interpretable than linear models \cite{lipton2016mythos}.  

In either case, if the ambiguities of the requirement of interpretability should render it ineffective in preventing higher order manipulations, what options remain available to regulators?  It may be that even without  comprehensive model understanding, reasonable guarantees about the actions taken by the model can be achieved.  In a number of complex industrial control systems, a policy function exists implicitly through a functional approximation to the physics of the system, rather than solely through the direct inference of system dynamics from data.  As an example, standard commercial autopilots rely on an implicit policy function through sophisticated control systems \citep{barros2011, apkarian1995self} and are required to provide reasonable guarantees about their behaviour to avoid undesirable outcomes for the operator.  To achieve this, controller designers choose approximations to the system dynamics in order to arrive at an implicit policy which is guaranteed to avoid unfavourable regions of state/action space.  There are a variety of subclasses of approximation which lead to provably \say{correct} systems (see \cite{isidori1995} for an example). While it remains challenging to provide guarantees on the long term behaviour of highly complex policy functions learned from data, there is growing interest in achieving accurate credible interval estimations for the outputs produced by deep neural networks \cite{Gal2016Dropout}. Research in this area may provide some empirical understanding of the behaviour we might expect from a given policy function as it adapts to new observations. Other methods have demonstrated the potential of combining a series of locally simple models \cite{2014-mfcgps}, an approach that has the potential to admit more accessible analysis.  If further work is able to provide appropriate state-action space behaviour guarantees, it should be able to restrict manipulative behaviour without a requirement on the low level interpretability of the policy function.

An interesting alternative for the regulation of algorithms that lie beyond human interpretation can be found in the growing field of \textit{machine ethics} \citep{moor2006nature, wallach2008moral}.  As the field of machine learning continues to develop, it may frequently be the case that the most useful algorithms do not readily admit human interpretation.  Rather than prohibiting the use of these algorithms, it may be possible for regulators to prevent manipulation by requiring that curator algorithms act in a manner that is consistent with a carefully specified set of ethical choices. However, we note that at present this approach faces a number of challenges that make the implementation of an ethical curator algorithm extremely difficult \cite{brundage2014limitations}.

While provable behavioural guarantees and machine ethics could prove to be effective tools for regulators in the long term, in the short term we suggest that a more pragmatic approach may be required.  As noted above, the third key factor enabling curator algorithms to develop manipulative behaviour is the availability of an immediate and effective feedback mechanism for evaluating the response of users.  We therefore suggest that a practical short term solution can be achieved through the construction a partial firewall restricting the flow of information that provides this mechanism in social networks.  However, if applied without careful consideration, this method runs the risk of placing an unnecessarily strong constraint on the ability of social network operators to improve their curation service for their users.  Regulators must therefore seek an appropriate balance between safeguarding users from the risks of manipulation and enabling operators to innovate and produce products which will benefit those users. 

As a brief aside, we note that here that although the financial industry differs in many ways from the world of social networking, it provides a useful reference for the difficult challenges facing regulators in the social network domain. Specifically, regulators seek to prevent \textit{market manipulation}, a practice by which participants artificially distort information to their benefit \cite{jarrow1992market}. However, while manipulation through human-based trading practices have long been outlawed, legislators are still exploring approaches to regulating the High Frequency Trading (HFT) algorithms that increasingly dominate the marketplace. By operating at speeds humans cannot match these algorithms are able to manipulate\footnote{One such technique is \textit{spoofing}, a practice which involves placing large sell orders above the current asking price which are quickly cancelled if the price begins to rise.} the market in ways that are difficult to detect \citep{muthuswamy2011high, lenglet2011conflicting}. The need for the regulation of these algorithms was brought into sharp relief by their role in the \say{Flash Crash} of the stock market in 2010, an event which resulted a $9\%$ index drop in a single hour of trading \cite{securities2010findings}.  One regulatory approach to preventing algorithmic market manipulation has been the introduction algorithm tagging, a process in which traders must provide the identity of the algorithm responsible for a trade \cite{germanHFT}. While this approach has been helpful in improving regulators' understanding of the interactions between different market participants, it has not yet been demonstrated to be effective in preventing manipulation \cite{coombs2016algorithm}.  At times, regulators have taken the more direct approach of requesting access to the algorithms themselves \cite{reuters}, but when algorithm interpretability is not feasible this action is of limited value.  In short, despite extensive experience in regulating trading practices to prevent market manipulation, financial industry regulators have yet to achieve a unified approach to the problem of algorithmic manipulation.  This should serve as a warning that a regulatory solution to the potentially more complicated issue of preventing human manipulation may prove extremely challenging.

In summary, the task facing regulators seeking to prevent the manipulation of users by social network curator algorithms is a difficult one.  The bold approach taking by the European Union may prove effective in combatting this issue, but it remains to be seen whether setting a requirement of interpretability is both practical and enforceable.  Should this be the case, we propose a simple firewall-based approach as short-term safeguard for users until more sophisticated techniques can be developed to prevent the risks of manipulation.

\section{Conclusions}

Machine learning applications have the potential to have an enormously positive impact on a wide range of industries and on the daily lives of people around the globe.  However, while it is encouraging to see the research and development of the algorithms driving these applications making rapid progress, it is important to note that when provided with access to abundant quantities of personal data, these algorithms also present risks.  In this work we discuss one such risk, namely the potential for the manipulation of users by the curator algorithm of a social network. To clarify the dangers associated with this possibility, we highlighted strategies we believe could be feasibly discovered through current reinforcement learning techniques given adequate access to available stored user data.  Regulators are faced with a difficult task as they try to allow new technologies to flourish while protecting the users of social networks from forms of manipulation that are inherently difficult to detect. In an effort to address this issue, we assessed potential avenues for regulating curator algorithms and offered recommendations for their use. As machine learning continues to progress, we expect to see similar tradeoffs between opportunities and risks emerge in across other industries which rely heavily on personal data.
\subsection{Acknowledgements}

The authors would like to thank Thomas Smith, Hannah Hjerpe-Schroeder, Ruth Fong, Ankush Gupta, James Thewlis, Anna-Elizabeth Shakespeare, Judith Albanie and Stephen Roberts for helpful discussions and Neil Lawrence, whose talks inspired much of this work. Samuel Albanie and Hillary Shakespeare are funded by the ESPRC EP/L015897/1 (AIMS CDT) grant. Tom Gunter is supported by UK Research Councils.

\medskip

\small

\bibliographystyle{unsrt}
\bibliography{references}

\end{document}